# Fused Deep Neural Network based Transfer Learning in Occluded Face Classification and Person re-Identification

Mohamed Mohana[1,*], Prasanalakshmi B [1,2], Salem Alelyani[1,3], Mohammed Saleh Alsaqer [1,3]

1 Center for Artificial Intelligence (CAI), King Khalid University, Saudi Arabia

2 College of Science and Arts, King Khalid University, Saudi Arabia

3 College of Computer Science, King Khalid University, Saudi Arabia

* mmuhanna@kku.edu.sa

**Abstract:**

Recent period of pandemic has brought person identification even with occluded face image a great importance with increased number of mask usage. This paper aims to recognize the occlusion of one of four types in face images. Various transfer learning methods were tested, and the results show that MobileNet V2 with Gated Recurrent Unit(GRU) performs better than any other Transfer Learning methods, with a perfect accuracy of 99% in classification of images as with or without occlusion and if with occlusion, then the type of occlusion. In parallel, identifying the Region of interest from the device captured image is done. This extracted Region of interest is utilised in face identification. Such a face identification process is done using the ResNet model with its Caffe implementation. To reduce the execution time, after the face occlusion type was recognized the person was searched to confirm their face image in the registered database. The face label of the person obtained from both simultaneous processes was verified for their matching score. If the matching score was above 90, the recognized label of the person was logged into a file with their name, type of mask, date, and time of recognition. MobileNetV2 is a lightweight framework which can also be used in embedded or IoT devices to perform real time detection and identification in suspicious areas of investigations using CCTV footages. When MobileNetV2 was combined with GRU, a reliable accuracy was obtained. The data provided in the paper belong to two categories, being either collected from Google Images for occlusion classification, face recognition, and facial landmarks, or collected in fieldwork. The motive behind this research is to identify and log person details which could serve surveillance activities in society-based e-governance.

## 1. Introduction:

Face masks have been made mandatory during those days of pandemic. This has hindered the identification of individuals in surveillance systems. Research has taken its path in identifying a person by their individual face features even upon wearing a medical face mask. Although the main focus of attention has been only on medical masks, there are other types of face occlusion, which also gain equal importance in the identification of a person. The research work has been carried out taking into consideration other types of occluded faces, initially being limited to four types to launch the face recognition model. The occlusions include medical masks, objects, scarves, hands. Occluded face recognition has become an outstanding problem in the domain of image processing and computer vision to favor e-governance systems. Face detection has more impact in face recognition where very high precision is preferred in terms of surveillance. In spite of a drastic development in the domain of computer vision and machine learning, a lot of issues related to occluded face detection are still to be addressed. This has become a high interest area for computer science, where the focus is not only on static images but also on video frames. Very high accuracy in image classification and object detection has already been achieved [1,2]. However, the detection of face occlusions is an extremely challenging task for the existing models of face detection [7,8,9,10]. Even though many methods have been proposed and many are in use for face detection and occlusion detection, challenges still exist when it comes to video surveillance. Problems occurring in existing systems are poor datasets and sometimes the existence of noise due to masks/occlusions. These issues have been studied using limited datasets [11,12,13], and the current challenge is to develop an efficient face occlusion detection model working for a vast dataset.

The model for face area detection proposed in this paper is done using OpenCV Deep Neural Network (DNN) [3], TensorFlow [4], Keras [5], and MobileNetV2 architecture [6], which is used as an image classifier. The proposed model can be integrated into surveillance systems for better person recognition, even under occlusion.

The main contributions of the paper include:

i. Real-time face occlusion detection through OpenCV DNN. Even faces under different orientations and occlusions can be detected and the proposed model outperforms the previous ones.
ii. Person identification with occluded faces.
iii. The fusion of RNN and DNN models to improve the accuracy of existing models. A state-of-the-art technique is proposed, namely MobileNetV2 [6] with GRU [14] component, to provide a precise classification.

The rest of the paper is structured as follows: Section 2 discusses the related work, detailing the recent technologies for face recognition and person identification; Section 3 discusses the proposed approach through a fusion between Deep neural network (MobileNetV2) and Recurrent Neural Network (GRU) accompanied with the ResNet backbone based Caffe implementation on face detection; Section 4 presents the results and their discussion, followed by a conclusion and future suggestions in Section 5.

## 2. Related Work

Face detection and identification have been studied by many researchers [15,16]. Existing works focus on face detection and occlusion classification. Face detection includes 20 detected facial components. AdaBoost based classifiers, Decision tree, and Linear Discriminant Analysis [17] have been used to classify non-occluded as well as occluded faces. El-Barkouky and colleagues [18] proposed a selective part model for partial occluded face detection. Gul and colleagues [19] used the Viola Jones approach [20] to detect partial faces forming the features used for face detection. Shape-based approaches [21,22,23,24,25] have also been used for face detection based on prior knowledge of head, neck, and shoulder shapes. Several approaches exist to solve occluded face detection issues [26-31]. The focus has been on gray scale face images as well [32]. In CNN-based classification, face detector models have been able to learn from users' data, then deep learning algorithms have been applied [33]. Cascade CNN models have been introduced [34]. For instance, AlexNet

Architecture has been upgraded [32] by fine-tuning the image dataset. CNN-based 3D models have been proposed [34] with the learning structure for face mask detection.

The proposed model has been developed using deep neural network modules from OpenCV and TensorFlow, which contains an object detection model. This was done using a face recognition [35] library. Typical classification architectures like ResNet-50, used as a backbone architecture, and image classification and fine-tuned fused MobileNetV2-GRU classifier have been used for this model. The fine-tuned fused MobileNetV2 architecture is made up of 17, $3 \times 3$ convolutional layers in a row accompanied by a $1 \times 1$ convolution, an average layer of max pooling, and some added layers for classification appended by GRU to improvise the classification accuracy. The residual connection is a new addition in the MobileNetV2 classifier. The MobileNetV2 model is computationally efficient for lightweight computational devices. Working with low-resolution images has motivated the authors to choose the MobileNetV2 model, and GRU is efficient in handling the gradient disappearing issue over the iterations in the neural networks that assist in faster training of the model. The proposed models adopt the strategy of a CNN backbone (MobileNet) and RNN units (GRU): the CNN backbone extracts the feature of images using predefined weights and uses the added layers in classification, fine-tuned using GRU through the softmax layer and the fully connected layer with the Rectified Linear Unit (ReLu), which connects the CNN backbone architecture with the RNN units.

3. **Proposed methodology**

To recognize a person with occlusions, initially, the model has to be trained using a proper dataset. Details about the Dataset will be discussed below in Section 3.1. This section describes the method of face identification using facial marks and bounding box to identify the region of interest using OpenCV, and explains the process involved in identifying the person based on the mask type using fused MobileNetV2-GRU. In the fused model, MobileNetV2 is used to classify the person and GRU enhances the performance.

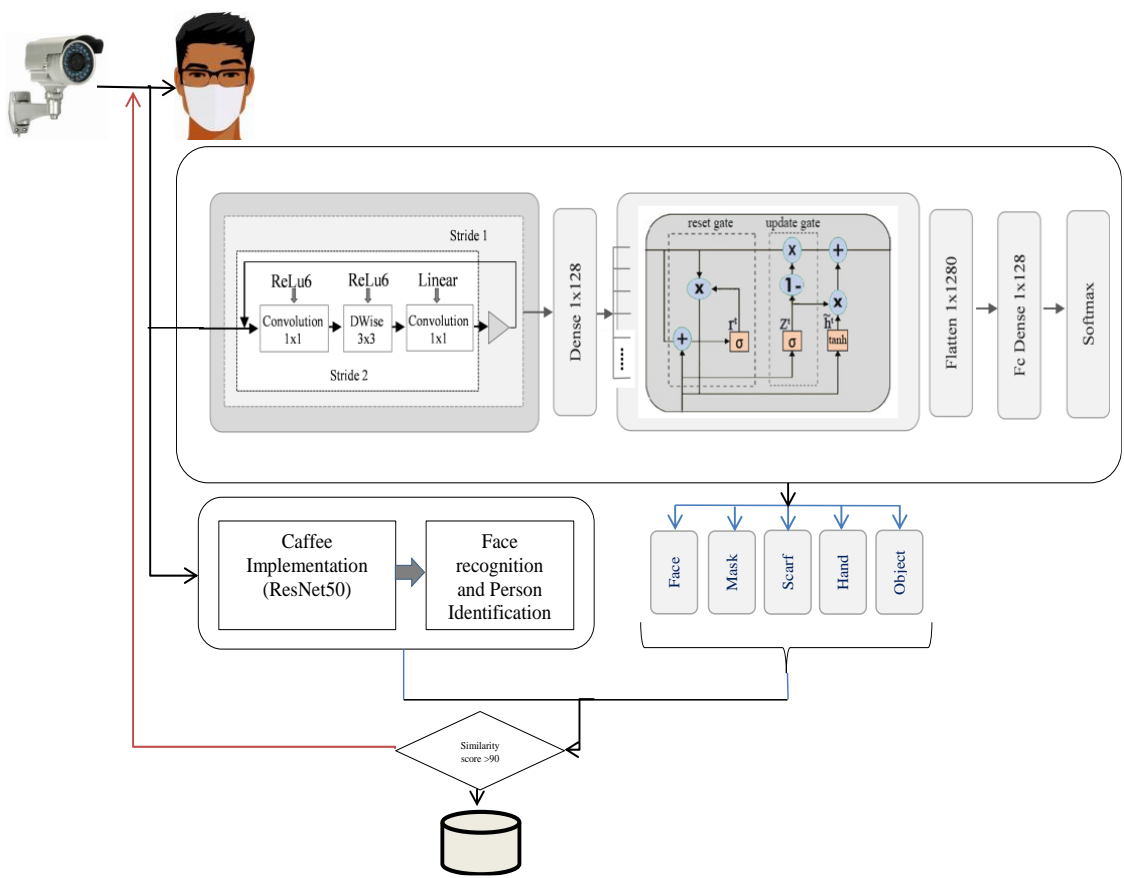

**Figure 1 Flowchart of Proposed model**

Initially, the face image captured by the video surveillance camera is dealt simultaneously for two-step processing. The first step involves the occlusion detection process through the fused MobilNetV2-GRU architecture, which classifies the obtained image based on the type of occlusion and identifies its corresponding face image. Since the dataset holds the names of persons as labels, the label of the image is identified. Simultaneously, the image captured from the video surveillance is dealt with the Caffe implementation of ResNet50 for identification of the person. Finally, the matching scores obtained from both stages are compared: if the matching score is above 90, the name of the person along with the type of mask and the time of data entry are logged in the database.

## 3.1 Dataset Collection and Preprocessing
a. Classification Dataset

To determine if a person is occluded or not and, if occluded, what the type of occlusion is among Hand, Scarf, Object, or Medical Mask categories, the images of such types of occlusions were taken from Google Images through a random search and manually labeled to facilitate the model to learn the type of occlusion.

Data sources include [36] and Kaggle datasets, that are publicly available.

Training Data

    |___ Name_Face(150 images+1918 images)
    |___ Name_ scarf(125 images)
    |___ Name_ Handocclusion(275 images)
    |___ Name_ Objectocclusion(260 images)
    |___ Name_ Medicalmask(1915 images+1918)

This dataset was used at stage 1 of the proposed model. The name of the person was identified from the label of the data.

b. Person Identification Dataset

The dataset was collected as shown in Fig 1. Individual images were collected for each specific person as five subfolders with real-time video frames saved as individual images. Each subfolder had 50 images, hence contributing to 250 images per person in five different categories of occlusion and forming a balanced dataset for each type of occlusion.

 Person1
    |___ Face
    |___ scarf
    |___ Handocclusion
    |___ Objectocclusion
    |___ Medicalmask

```
Person2
   |___ Face
   |___ scarf
   |___ Handocclusion
   |___ Objectocclusion
   |___ Medicalmask
```

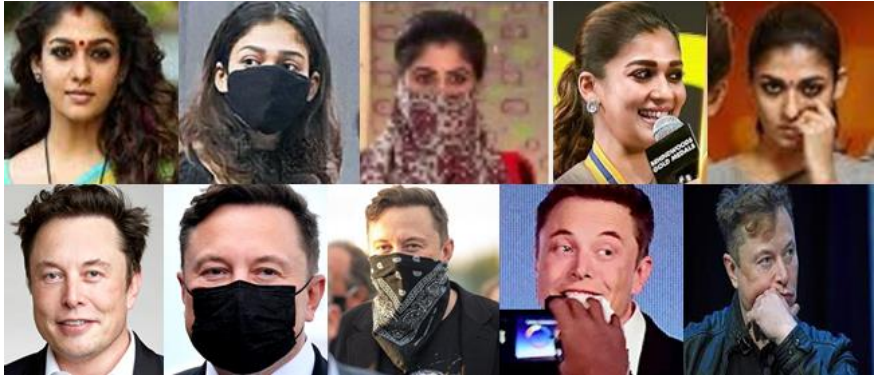

**Fig 2 .Sample images.**

### 3.2    Data Augmentation

Deep learning models have a high learning capacity that allows them to solve classification and prediction tasks with relevant results, particularly on perceptual problems that receive high-dimensional samples as input images. However, complex models tend to display a decreased generalization capability when trained with a small dataset, an issue known as overfitting. Data augmentation is a technique adopted to mitigate overfitting in computer vision. It takes the approach of generating more training data from existing training samples, by augmenting the samples via a number of random transformations that yield believable-looking images (Fig. 3). In the literature, several methods are adopted to introduce more variability into the dataset: standard techniques include rotation, shearing, zooming, cropping, flipping, and fill. In this work, we relied on standard augmentation to improve the performance and generalization ability of the model.

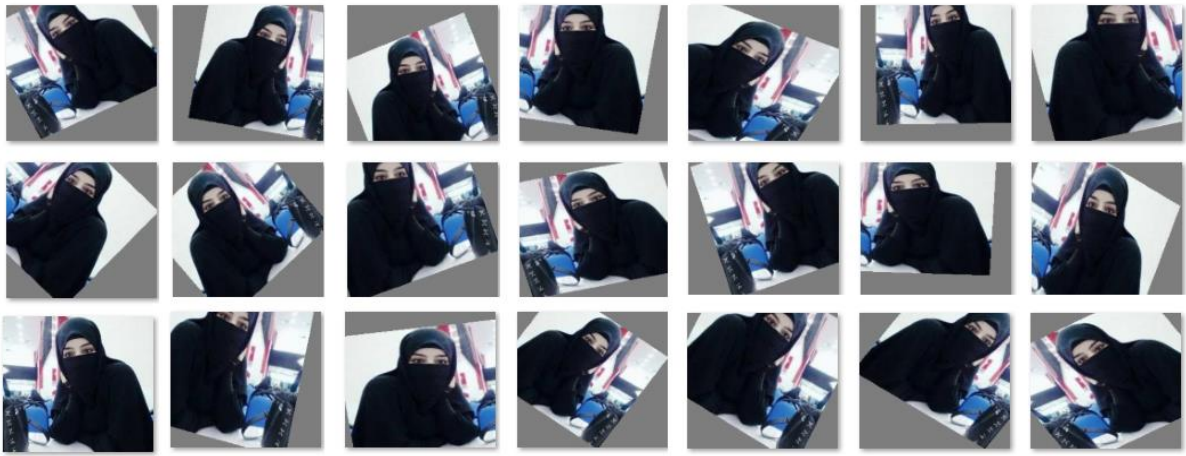

Figure 3 Augmented Sample Images

During the training, a huge number of images will be available, as each and every augmentation will have been applied to every image. All images have been resized to 224 x 224. Asian, European, Indian, Arab people, women wearing hijab, men with beards, aged people, almost all types of masks are included within the dataset. The images of data augmentation counts up to 20 images per category per person, with a total of 5000 images of a person under five different conditions of occlusion.

### 3.3  Face Detection using OpenCV DNN

The Caffe implementation of ResNet50 is taken into consideration for face recognition. Caffe is a deep learning framework to work faster, powerful and efficient on object detection. The backbone as selected to be ResNet 50 reason out that it increases the accuracy when compared to ResNet10 or ResNet18, even though the time taken for computation is little varying which could be compromised. ResNet with increased number of layers proves almost the same accuracy by the execution time varies a lot which insisted the choose ResNet 50. Table 1 shows the variation as explained earlier. Testing accuracy and prediction time were done in Keras implementation which when tested on Caffe implementation would be more precise. In order to justify the selection method the implementation in keras has been done , definitely which outperforms in OpenCV implementations.

**Table 1. Comparison Results of different ResNet models on created dataset**

| Models | Training Accuracy | Testing Accuracy | Prediction Time |
|---|---|---|---|
| ResNet10 | 0.894 | 0.80 | 1028s |
| ResNet18 | 0.969 | 0.85 | 1133s |
| ResNet50 | 0.982 | 0.92 | 1412s |
| ResNet101 | 0.9819 | 0.96 | 1821s |
| ResNet152 | 0.987 | 0.96 | 2256s |

ResNet displayed almost the same accuracy by execution time with an increased number of layer and ResNet50 was chosen on this basis. Table 1 shows the variation as explained earlier. Testing accuracy and prediction time were done in Keras implementation which, when tested on Caffe implementation, would be more precise. In order to justify the selection method, the implementation in Keras has been done, definitely outperforms in OpenCV implementations.

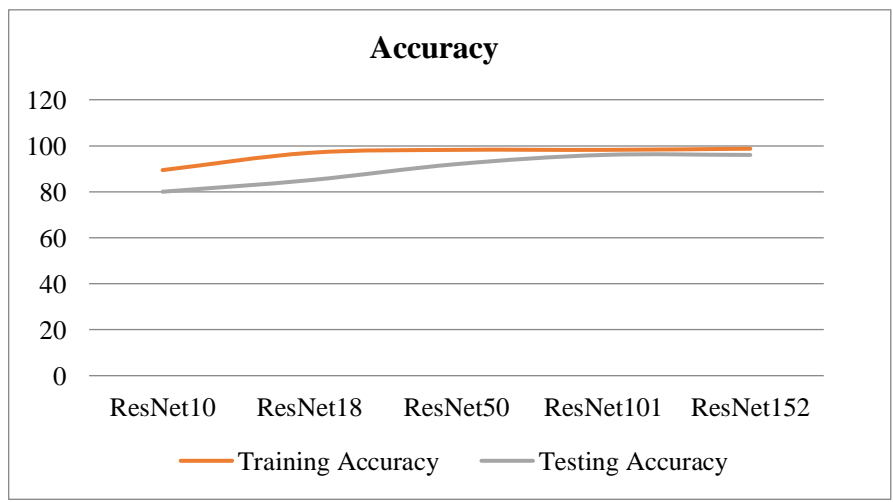

**Figure 4. Comparison of Testing and Training accuracies**

Since the accuracy and prediction time favored the selection of ResNet50, its Caffe implementation was taken as implementation criteria. Faces of different orientations, size, age, gender were detected with better accuracy.

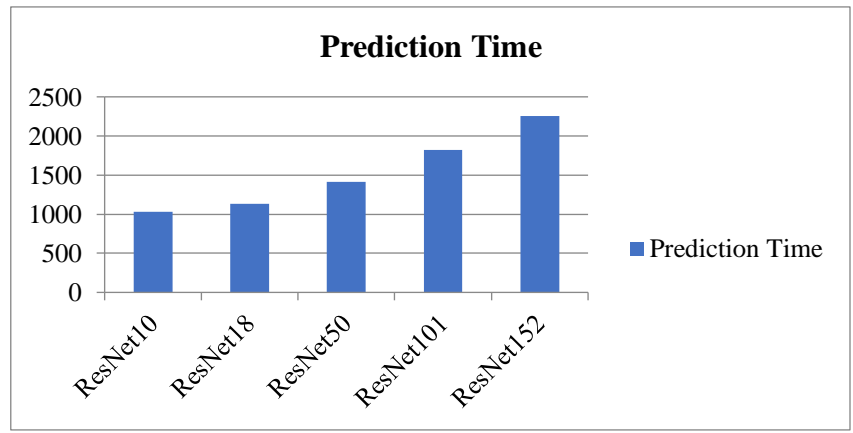

**Figure 5 Prediction time comparison**

These outputs were then given as input for the person identifier. Hence, the Caffe implementation not only detects the face but also helps the person identifier to locate an already registered similar face image.

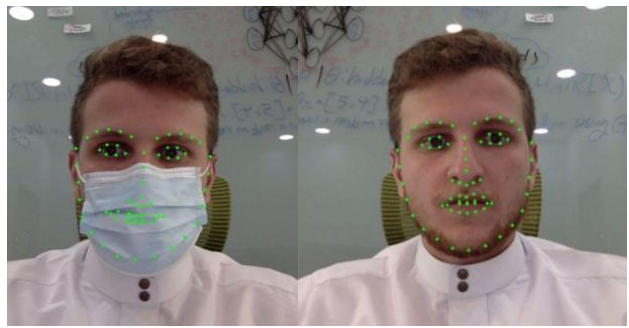

**Figure 6 Facial Landmark detection**

The detected face is selected for preliminary identification which seeks for a matching score in the re-identification process, the final step of the prototype.

### 3.4 Fine-tuned MobileNet V2

The next stage involves the observation of the detected face, if the response turns positive the captured face images are fed to the MobileNet architecture for further analysis.

When compared to a conventional CNN based model MobileNet V2 [6] is widely used for image classification. Its main advantage is the less computational effort which enables it to work in mobile devices and Internal storage enabled IoT devices like surveillance cameras.

MobileNetV2 is an updated version of MobileNetV1 that does not use of any special operators while improving accuracy, and achieving the state of the art on multiple image classification and detection tasks suitable for mobile applications. MobileNetV2 is a depth-wise separable convolution neural network. The basic idea is the use of two separate convolution layers namely a depth-wise convolution layer to facilitate light-weight filtering by applying a single convolution filter per input channel and a 1 × 1 convolution responsible for building new features through linear combination computation of input channels. The input is converted into tensors $\tau_i$ of size $h_i \times w_i$ pixels of $d_i$ dimensions and the output is obtained as a tensor $\tau_j$ of size $h_j \times w_j$ pixels of $d_j$ dimensions. The computation cost of the depth-wise separable convolution is the sum of cost of depth-wise convolution and the 1 × 1 pointwise convolution, represented as,

$$Cost = h_i \cdot w_i \cdot d_i \cdot k^2 + h_i \cdot w_i \cdot d_i \cdot d_j \dots\dots(1)$$

Where $k$ applies to convolutional kernel. To reduce the space complexity the dimensionality reduction of layer is done by adding the width multiplier layer, normalised to 1, when ReLu is used as the activation function. Hence the depletion variable D, is approximated as,

$$D = \frac{h_i \cdot w_i \cdot d_i \cdot k^2 + h_i \cdot w_i \cdot d_i \cdot d_j}{h_i \cdot w_i \cdot d_i \cdot d_j \cdot k^2} \dots\dots(2)$$

The proposed model uses the input image of 224 ×224 ×3 where $h_i \times w_i$ is 224 ×224 and $d_i$ is 3.

The building blocks of MobileNetV2 includes:

- 1 ×1 convolution with ReLu6 in the first layer
- Depth-wise 3 ×3 convolution with ReLu6.
- 1 ×1 linear convolution layer
- Residual component to support the gradient flow across the network through batch processing and the activation function ReLu6.

### 3.5 MobileNetV2 with BPTT trained GRU

When flowing through the network, the gradient may shrink leading to a vanishing gradient problem, thus decaying the performance and accuracy. GRU is like LSTM having its ability to model longer sequences. While LSTM has three gates, GRU has only two: an update and a reset gate. A standard backpropagation training algorithm (BPTT) optimizes the weights based on the resulting network output error. To overcome the vanishing gradient problem, GRU uses BPTT to compute gradients.

The key idea of GRUs is that the gradient chains do not vanish due to the length of sequences. This is done by allowing the model to pass values completely through the cells. The model is defined by the following set of equations:

$$h_t = f \odot h_{t-1} + (1-f) \odot g' \dots(3)$$

$$g' = \tanh(W_g(r \odot h_{t-1}) + U_g x_t + b_g) \dots(4)$$

$$f = \sigma(W_f h_{t-1} + U_f x_t + b_f) \dots(5)$$

$$r = \sigma(W_r h_{t-1} + U_r x_t + b_r) \dots(6)$$

$$\hat{y} = softmax(V h_t + b) \dots(7)$$

The input to hidden connections is parametrized by a weight matrix U, hidden-to-hidden recurrent connections by a weight matrix W and hidden-to-output by weight matrix V with a bias vectors b. In the definitions, ∘ represents the Hadamard product, for element-wise multiplication; σ(x) is the Sigmoid function that has the effect of squishing values between 0 and 1 – useful in updating or forgetting data as any number multiplied by 0 is 0, causing values to disappears or be "forgotten". Equally, multiplying a number by 1 retains the value hence propagating it forward into other layers in the network. A tan-h function ensures that the values stay between -1 and 1, thus regulating the output of the neural network. f functions as a filter for the previous state. If it is low (near 0), then a substantial amount of the previous state value is reused. The size of the input vectors at the current state $x_t$ has less influence on the output. If f is high, then the output at the current step is largely influenced by the current input $x_t$, and less by the size of the hidden layer vectors from the previous state $h_{t-1}$. r functions as forget gate (or reset gate), allowing the cell to forget certain information of the state.

The categorical cross entropy loss is used for the multi classification and the loss function is defined as

$$L = \sum_t L_t(y_t - \hat{y}_t) = -\sum_t y_t \, log\hat{y}_t \dots\dots(8)$$

The process of truncated BPTT calculates the gradient using the following formulation procedure:

$$\frac{\partial L_t}{\partial V} = (\hat{y}_t - y_t) \otimes h_t \dots\dots(9)$$

$$\frac{\partial L_t}{\partial W} = \sum_t \frac{\partial L_t}{\partial \hat{y}_t} \frac{\partial \hat{y}_t}{\partial h_t} \left( \prod_{j=k+1}^{t} \frac{\partial h_j}{\partial h_{j-1}} \right) \frac{\partial h_k}{\partial W} \dots\dots(10)$$

$$\frac{\partial L_t}{\partial U} = \sum_t \frac{\partial L_t}{\partial \hat{y}_t} \frac{\partial \hat{y}_t}{\partial h_t} \left( \prod_{j=k+1}^{t} \frac{\partial h_j}{\partial h_{j-1}} \right) \frac{\partial h_k}{\partial U} \dots\dots(11)$$

Hence, the gradients are calculated using formulas 9-11 to decide on the weight adjustments to obtain a perfect output.

The output of the fine-tuned MobileNetV2-GRU fused architecture is the classified result, identifying the category of occlusion to which the image belongs. The purpose of identifying the occlusion is for further investigations or to backtrack history in the field of investigation on surveillance. The category that is obtained as output of the fused architecture serves as an input for the next stage of face recognition which uses the defined face recognition library from Dlib for the identification.

### 3.6 Face recognition

For the process of face recognition, images with several occlusions were collected from each person and stored with the person's name. Face recognition library [5] is used to recognize and identify the occluded face of the person and register their names as a csv file with their name, type of occlusion, and time. An example of stored data log is shown in Figure 7.

| Date | Time | Person | occlusion | type |
|---|---|---|---|---|
| 25-Jun-21 | 10:30 AM | Mohamed | Yes | Medical |
| 25-Jun-21 | 11:28 AM | Vijay | No | NA |
| 26-Jun-21 | 12:30 PM | Mohamed | Yes | object |
| 27-Jun-21 | 1:30 PM | Vijay | Yes | scarf |

**Figure 7. excel log database**

### 4. Results and discussion

In this section, the results of the proposed model are discussed in detail. The proposed MobileNetV2 with GRU performance is evaluated through the hyperparameters like training and validation loss measures that determine the proposed model's capabilities. The learning rate at various training levels is discussed in the current section. The performance evaluation with other existing approaches in terms of Sensitivity, Specificity, Accuracy, Jaccard Similarity Index (JSI), and Mathew Coefficient Correlation (MCC) are presented. The proposed model's computational time is evaluated as a part of performance evaluation and compared with the existing approaches on performing the classification over similar data. Since the proposed model shows an emphasis on the fused architecture, the result evaluation of it is well-defined in this section.

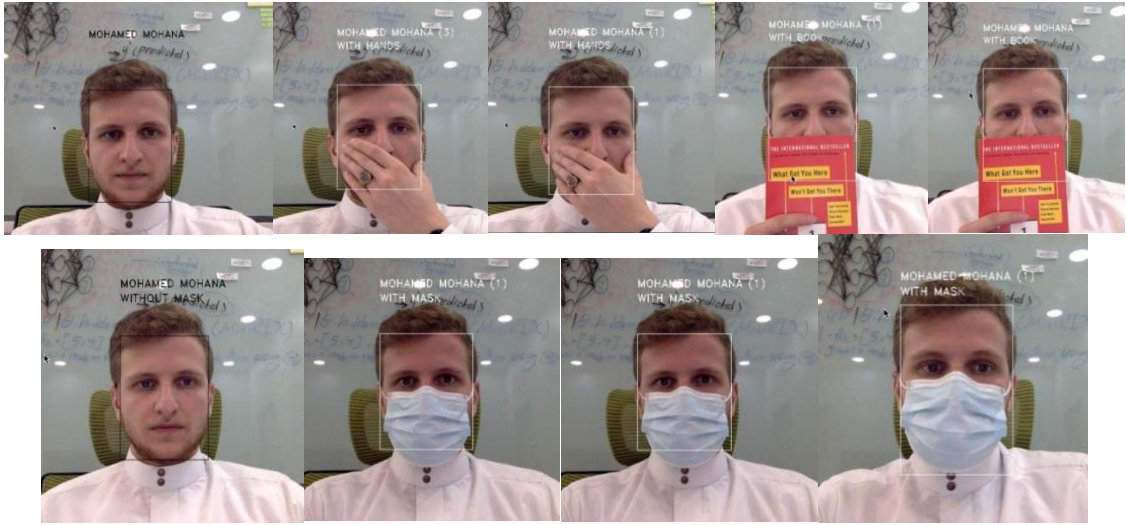

**Figure 8 Occlusion Detection**

### 4.1 Performance Evaluation of the proposed model

The proposed model was implemented in the dataset described in section 3.1. The results show the classification accuracy for each test dataset.

To make a reasonable justification the implementation configuration are standardized throughout the implementation and result evaluation. The parameters considered in the implementation of the proposed model are mentioned below.

*Implementation configuration parameters*

*Model: MobileNetV2*

*Base learning rate :0.1*

*Learning rate policy: stepwise (Reduced by a factor of 10 every 30/3 per epoch*

*Momentum:0.95*

*Weight Decay:0.0001*

*Cycle length:10*

*PCT-Start :0.9*

*Batch Size : 50*

*Optimiser : Adam*

Preliminarily, the experiment was performed over several images and the type of occlusion was assessed through the proposed fused MobileNetV2-GRU approach. The graphs represented in Figure 9 were obtained from the initial trained model, where the training model loss is better than the validation loss. The left graph indicates the number of batches processed versus loss obtained during the training and the validation phases. The batch size value in the initial model is 100, which is used to speed up the training data.

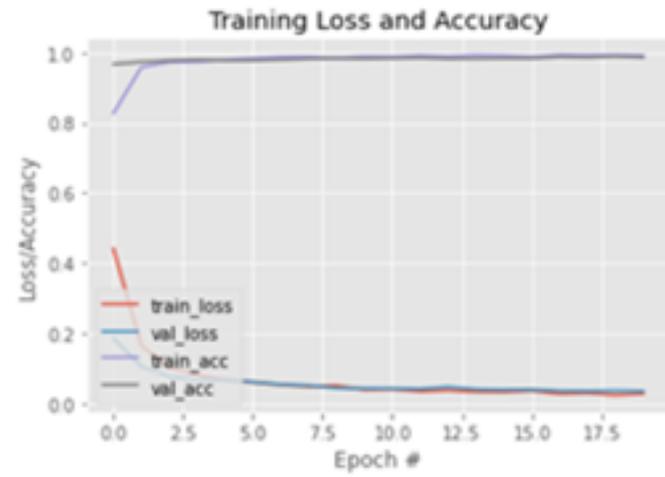

Figure 9 Training loss and accuracy

The training and validation loss alongside the learning rate are presented in Figure 9, and they are significant in determining the overfitting and underfitting of the proposed model. When the validation loss is ahead of the training loss, the model may end up overfitting; when they are almost equal, there would be an underfitting problem.

Figure 6 with graphs and outputs is observed by the trained model before improvements of the training data, and Figure 8 presents the results that were obtained by the trained model after slight improvements in terms of epochs, batch size, and data augmentation values. The batch size was reduced from 100 to 50 to reduce the computational time and also overcome the lower generalization results and higher loss values. The epochs value was increased by 20 to gain more accuracy. Data augmentation was also performed to reduce overfitting while training and minimizing the error rate. The batch size was kept more for speedup of the previous model's training data, which ended up getting lower generalization results.

Table 2. Performance metrics compared

| Model | Accuracy | f1 with Occlusion | f1 without Occlusion | Precision with Occlusion | Precision without Occlusion | Recall with Occlusion | Recall without Occlusion |
|---|---|---|---|---|---|---|---|
| MobileNetV2+GRU | 99.9 | 99.9 | 99 | 98 | 99 | 99 | 98 |
| MobileNetV2 | 99.1 | 99.1 | 99 | 98 | 99 | 99 | 98 |
| VGG16 | 96 | 96 | 96 | 95 | 97 | 97 | 95 |
| VGG19 | 96 | 96 | 96 | 97 | 96 | 96 | 97 |
| ResNet101 | 87 | 88 | 87 | 87 | 88 | 89 | 86 |
| ResNet50 | 83 | 81 | 84 | 91 | 77 | 73 | 92 |
| ResNet152 | 79 | 77 | 81 | 87 | 74 | 69 | 89 |
| MobileNetV3Large | 75 | 77 | 72 | 81 | 70 | 64 | 85 |
| MobileNetV3Small | 72 | 71 | 74 | 75 | 70 | 66 | 78 |
| EfficientNetB7 | 50 | 67 | 0 | 50 | 0 | 10 | 0 |

The graph represents the loss values versus processed batches, displaying higher loss values compared to the improvised model. The learning rate of the previous model is low compared to the final model. The learning rate is the hyperparameter that determines the weight of the network component. If the learning rate is too low, it becomes a challenging task and can also lead the process to get stuck. To overcome the drawbacks mentioned above, we reduced the batch size to a much smaller size to have faster convergence, resulting in optimized results. We increased the learning rate, leading to better outputs at training upon fewer epochs. The proposed model's value was assessed through various performance evaluation metrics like Sensitivity, Specificity, Accuracy, JSI, and the MCC. The models mentioned above were assessed through the True Positive, True Negative, False Positive, and False Negative values assessed through the repeated experimentation of the proposed approach. True Positive concerns the precise identification of the region of occlusion; True Negative represents the preciseness of the non-occluded region of the face, evaluated from the captured image; False Positive represents the number of times the proposed approach fails in recognizing the class of occlusion accurately; and False Negative determines the number of times the proposed model misinterprets a non-occluded region as an occluded region.

Figures 6 and 8 are the resultant hyperparameter graphs obtained upon the execution of the proposed model. Both graphs show that the training and the validation loss curves are close to each other, which depicts an optimal classification of the occlusion. The learning curve presents the reasonable level of the learning aspect of the model.

In evaluating the proposed model's performance, the experimentation was repeatedly executed over the auxiliary computer. The evaluations were done considering the number of times the proposed model accurately classifies the occlusion as True Positive and correctly identifies that the image is not of that particular occlusion category as True Negative. The number of times the proposed model recognizes the occlusion type correctly was considered a False Positive. The number of times the proposed model misinterprets the occlusion is assumed as a False Negative. The approximated True Positive (TP), True Negative (TN), False Positive (FP), and False Negative (FN) values were considered for evaluating metrics like Sensitivity, Specificity, and Accuracy of the proposed model. The values of the various evaluation metrics like Sensitivity, Specificity, Accuracy, Jaccard Similarity Index, and Matthews Correlation Coefficient are presented through Equations 12-16 with respect to the obtained True Positive, True Negative, False Positive, and False Negative values upon experimentation. The metrics determine the preciseness of the model in correctly classifying the occlusion.

$$Sensitivity = \frac{TP}{TP+FN} \ldots(12)$$

$$Specificity = \frac{TN}{FP+TN} \ldots(13)$$

$$Accuracy = \frac{TP+TN}{TP+FP+TN+FN} \ldots(14)$$

$$\text{Jaccard Similarity Index} = \frac{TP}{TP+FP+TN+FN} \ldots(15)$$

$$\text{Matthews Correlation Coefficient} = \frac{(TP \times TN)-(FP \times FN)}{\sqrt{(TP+FP)X\ (TP+FP)X\ (TN+FP)X\ (TN+FN)}} \ldots(16)$$

Table 3 reports the performance of our proposed approach and other related approaches in terms of Sensitivity, Specificity, Accuracy, JSI, and MCC.

**Table 3 Occlusion type based average accuracy**

| Occlusion type | Accuracy | JSI | MCC |
|---|---|---|---|
| Face | 99.89 | 93 | 88 |
| Medical Mask | 99.12 | 91 | 86 |
| Scarf / Niqab | 96.1 | 82 | 79 |
| Hand | 99.02 | 88 | 84 |
| Object | 98.6 | 81 | 80 |

MobileNet-based models exhibited a better performance in classifying the Region of interest with minimal computational efforts; MobileNetV2 exhibited an optimal efficiency in occlusion classification. The MobileNetV2 model encompassed GRU having an impact on crucial parameters such as learning rates and input and output gates, yielding a better outcome. Plotting the results of Table 2 in Figure 9, it is visible that the proposed MobileNetV2-GRU approach outperformed other state-of-the-art models in almost all performance sectors.

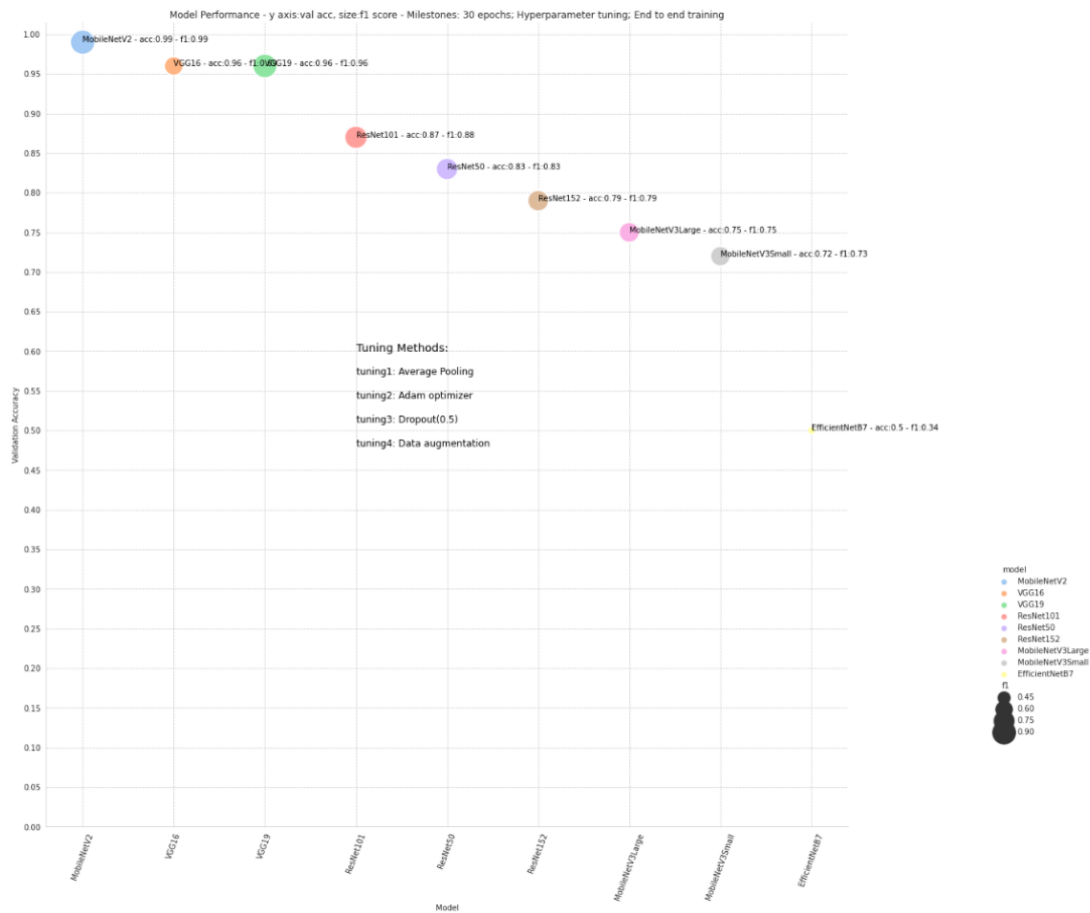

Figure 10 Model Performance and hyperparameter tuning

The incorporation of GRU component has enhanced the accuracy of the proposed approach. It can be observed from Table 3 that the proposed MobileNetV2 with GRU model has outperformed other tested models in terms of Sensitivity, Specificity, Accuracy metrics as well as the MCC and JSI [37-39].

### 4.3. Execution Time

In the process of evaluating the performance of the proposed model, the execution time of the validation phase is presented in Table 4 and Figure 13 in accordance with existing studies. The proposed model required approximately around 1134 seconds for training the model over 20 epochs. The computational time to MobileNetV2 with GRU over MobileNetV2 was not drastically reduced [40,41]. Still, MobileNetV2 exhibited a better prediction accuracy in terms of other performance evolution parameters like Sensitivity, Specificity, and Accuracy. The computational time of the proposed MobileNetV2 with GRU is reasonably good, as shown in Table 4, which makes it feasible to incorporate the technology to run over the computationally lightweight devices. Incorporating the GRU module will assist in faster convergence by remembering the significant features necessary for a more rapid and accurate classification of the occlusion.

**Table 4 Execution time compared**

| Algorithm | Execution Time(s) |
|---|---|
| MobileNetV2+GRU | 99.21 |
| MobileNetV2 | 98.92 |
| VGG16 | 106.54 |
| VGG19 | 110.26 |
| ResNet101 | 113.35 |
| ResNet50 | 115.86 |
| ResNet152 | 123.54 |
| MobileNetV3Large | 134.76 |
| MobileNetV3Small | 144.81 |
| EfficientNetB7 | 152.91 |

### 5. Conclusions

The proposed fused MobileNetV2 -GRU approach proved efficient for occlusion detection and face recognition with minimal computational power and effort. The outcome is promising, with an average accuracy of 99% in experiments compared with other methods over the real time images. The MobileNetV2 architecture is designed to work with a portable device with a stride2 mechanism. The model is computationally effective and the use of the GRU module with the MobileNetV2 would enhance the prediction accuracy by maintaining the previous timestamp data. The information related to the current state through weight optimizations would make the model robust. This was also compared with various other conventional models: we observed that the proposed model has outperformed them in face identification and classification, as discussed in the previous section. However, there is currently a range of shortcomings that must be resolved in future work. The model's precision is dramatically decreased to just below 80% when checked on a series of photographs captured in poor illumination conditions distinct from those used during testing. The proposed model is computationally efficient as it is designed to work on top of lightweight capability devices. Alongside the bottleneck in residual connections in the proposed architecture, the model yields higher accuracy with minimal effort. The model can be further improved by incorporating the self-learning capability and knowledge acquisition from its previous experiences. The efforts on training the model can be considerably reduced. However, the model must be mechanized to assess the impact of features extracted for each strategy, and the incorporation of randomizing components is necessary.

## 6. Future Works

The proposed model outperforms the other selected models in terms of several parameters. However, it was tested on limited datasets collected by means of search terms on search engines. Only a limited set of personally collected real time data were used in the study. This could be extended in future with more data on real time occlusion collected on different cultural and demographic bases. Another limitation includes the implementation in multiple IoT platforms. The considered types of occlusion were only four, which could be generalized and extended to as many occlusions as possible, such as sunglasses, caps, and others. The distance covered on capturing the real time face image was restricted to 2-3 m, which could be extended for testing in the future.

## References

[1] Ahmed, I., Ahmad, M., Rodrigues, J. J., Jeon, G., & Din, S. (2020). A deep learning-based social distance monitoring framework for COVID-19. Sustainable Cities and Society, Article 102571.

[2] Lawrence, S., Giles, C. L., Tsoi, A. C., & Back, A. D. (1997). Face recognition: A convolutional neural-network approach. IEEE Transactions on Neural Networks, 8(1),98–113.

[3] Velasco-Montero, D., Fern´andez-Berni, J., Carmona-Galan, ´R., & Rodríguez-Vazquez, ´A. ´(2018). Performance analysis of real-time DNN inference on Raspberry Pi. May. Real-Time Image and Video Processing 2018 (Vol. 10670, p. 106700F). International Society for Optics and Photonics.

[4] Abadi, M., Agarwal, A., Barham, P., Brevdo, E., Chen, Z., Citro, C., … Ghemawat, S.(2016). Tensorflow: Large-scale machine learning on heterogeneous distributed systems. arXiv preprint arXiv:1603.04467.

[5] face recognition library https://pypi.org/project/face-recognition/

[6] Nguyen, h. (2020). Fast object detection framework based on mobilenetv2 architecture and enhanced feature pyramid. Journal of Theoretical and Applied Information Technology, 98(05).

[7] Chen, D., Hua, G., Wen, F., & Sun, J. (2016). Supervised transformer network for efficient face detection. October. European conference on computer vision (pp. 122–138). Cham: Springer.

[8] Ranjan, R., Patel, V. M., & Chellappa, R. (2017). Hyperface: A deep multi-task learning framework for face detection, landmark localization, pose estimation, and gender recognition. IEEE Transactions on Pattern Analysis and Machine Intelligence, 41(1), 121–135.

[9]Zhang, K., Zhang, Z., Li, Z., &Qiao, Y. (2016a). Joint face detection and alignment using

multi-task cascaded convolutional networks. IEEE Signal Processing Letters, 23(10), 1499–1503.

[10] Zhu, C., Zheng, Y., Luu, K., &Savvides, M. (2017). Cms-rcnn: Contextual multi-scale region-based cnn for unconstrained face detection. Deep learning for biometrics (pp. 57–79). Cham: Springer.

[11] Ghiasi, G., & Fowlkes, C. C. (2014). Occlusion coherence: Localizing occluded faces with a hierarchical deformable part model. Proceedings of the IEEE conference on computer vision and pattern recognition, 2385–2392.

[12] Opitz, M., Waltner, G., Poier, G., Possegger, H., & Bischof, H. (2016). Grid loss: Detecting occluded faces. October. European conference on computer vision (pp. 386–402). Cham: Springer.

[13] Yang, S., Luo, P., Loy, C. C., & Tang, X. (2015). From facial parts responses to facedetection: A deep learning approach. Proceedings of the IEEE International Conference on Computer Vision, 3676–3684.

[14] Cho, Kyunghyun; van Merrienboer, Bart; Gulcehre, Caglar; Bahdanau, Dzmitry; Bougares, Fethi; Schwenk, Holger; Bengio, Yoshua (2014). "Learning Phrase Representations using RNN Encoder-Decoder for Statistical Machine Translation". arXiv:1406.1078.

[15] S. Liao, A. K. Jain and S. Z. Li, Partial face recognition: Alignment-free approach, IEEE Trans. Pattern Anal. Mach. Intell. 35 (2013) 1193–1205.

[16] J. Shermina and V. Vasudevan, Recognition of the face images with occlusion and expression, Int. J. Pattern Recogn. Artif. Intell. 26 (2012) 1256006-1–1256006-26.

[17] K. Ichikawa, T. Mita, O. Hori and T. Kobayashi, Component-based face detection method for various types of occluded faces, 3rd Int. Symp. Communications, Control and Signal Processing (2008), pp. 538–543.

[18] A. El-Barkouky, A. Shalaby, A. Mahmoud and A. Farag, Selective part models for detecting partially occluded faces in the wild, 2014 IEEE Int. Conf. Image Processing (ICIP) (2014), pp. 268–272.

[19] S. Gul and H. Farooq, A machine learning approach to detect occluded faces in unconstrained crowd scene, 2015 IEEE 14th Int. Conf. Cognitive Informatics Cognitive Computing (2015), pp. 149–155.

[20] P. Viola and M. Jones, Rapid object detection using a boosted cascade of simple features, Computer Visi. Pattern Recogn. 1 (2001) 511–518.

[21] T. Charoenpong, Face occlusion detection by using ellipse fitting and skin color ratio, Burapha University International Conf. (BUU), (2013), pp. 1145–1151.

[22] T. Charoenpong, C. Nuthong and U. Watchareeruetai, A new method for occluded face detection from single viewpoint of head, 2014 11th Int. Conf. Electrical Engineering/ Electronics, Computer, Telecommunications, and Information Technology (ECTI-CON) (2014), pp. 1–5.

[23] S. Hongxing, W. Jiagi, S. Peng and Z. Xiaoyang, Facial area forecast, and occluded face detection based on the ycbcr elliptical model, Int. Conf. Mechatronic Sciences, Electric Engineering and Computer (MEC) (2013), pp. 1199–1202.

[24] G. Kim, J. K. Suhr, H. G. Jung and J. Kim, Face occlusion detection by using b-spline active contour and skin color information, 11th Int. Conf. Control Automation Robotics Vision (ICARCV) (2010), pp. 627–632.

[25] D.-T. Lin and M.-J. Liu, Face occlusion detection for automated teller machine surveillance, in Advances in Image and Video Technology, Lecture Notes in Computer Science, Vol. 4319 (Springer Berlin Heidelberg, 2006), pp. 641–651.

[26] J. Kim, Y. Sung, S. Yoon and B. Park, A new video surveillance system employing occluded face detection, in Innovations in Applied Artificial Intelligence (Springer Berlin Heidelberg, 2005), pp. 65–68.

[27] C. Szegedy, S. Reed, D. Erhan and D. Anguelov, Scalable, high-quality object detection, arXiv preprint (2014).

[28] R. Girshick, J. Donahue, T. Darrell and J. Malik, Rich feature hierarchies for accurate object detection and semantic segmentation, 2014 IEEE Conf. Computer Vision and Pattern Recognition (CVPR) (2014), pp. 580–587.

[29] K. He, X. Zhang, S. Ren and J. Sun, Spatial pyramid pooling in deep convolutional networks for visual recognition, IEEE Trans. Pattern Anal. Mach. Intell. 37 (2015) 1904–1916.

[30] X. Wang, M. Yang, S. Zhu and Y. Lin, Regionlets for generic object detection, IEEE International Conference on Comput. Vis. (ICCV) (2013), pp. 17–24.

[31] P. Felzenszwalb, R. Girshick, D. McAllester and D. Ramanan, Object detection with discriminatively trained part-based models, IEEE Trans. Pattern Anal. Mach. Intell. 32 (2010) 1627–1645.

[32] Ojala, T., Pietikainen, M., &Maenpaa, T. (2002). Multiresolution gray-scale and rotation invariant texture classification with local binary patterns. IEEE Transactions on Pattern Analysis and Machine Intelligence, 24(7), 971–987.

[33] Ren, S., He, K., Girshick, R., & Sun, J. (2015). Faster R-CNN: Towards real-time object detection with region proposal networks. Advances in neural information processing systems (pp. 91–99).

[34] Li, H., Lin, Z., Shen, X., Brandt, J., & Hua, G. (2015). A convolutional neural network cascade for face detection. Proceedings of the IEEE conference on computer vision and pattern recognition, 5325–5334.

[35] https://pypi.org/project/face-recognition/

[36] Zhongyuan Wang, Guangcheng Wang, Baojin Huang, Zhangyang Xiong, Qi Hong, Hao Wu, Peng Yi, Kui Jiang, Nanxi Wang, Yingjiao Pei, Heling Chen, Yu Miao, Zhibing Huang, Jinbi Liang, Masked face recognition dataset and application, 2020.

[37] He, K.; Zhang, X.; Ren, S.; Sun, J. Deep residual learning for image recognition. In Proceedings of the 2016 IEEE Conference on Computer Vision and Pattern Recognition (CVPR), Las Vegas, NV, USA, 27–30 June 2016; pp. 770–778.

[38] Mahdianpari, M.; Salehi, B.; Rezaee, M.; Mohammadimanesh, F.; Zhang, Y. Very Deep Convolutional Neural Networks for Complex Land Cover Mapping Using Multispectral Remote Sensing Imagery. Remote Sens. 2018, 10, 1119.

[39] Songtao, G.; Zhouwang, Y. Multi-Channel-ResNet: An integration framework towards skin lesion analysis. Inform. Med. Unlocked 2018, 12, 67–74

[40] Xiang, Q.; Wang, X.; Li, R.; Zhang, G.; Lai, J.; Hu, Q. Fruit Image Classification Based on MobileNetV2 with Transfer Learning Technique. CSAE 2019. Proceedings of the 3rd International Conference on Computer Science and Application Engineering, Sanya, China, 22–24 October 2019; pp. 1–7.

[41] He, D.; Yao, Z.; Jiang, Z.; Chen, Y.; Deng, J.; Xiang, W. Detection of Foreign Matter on High-Speed Train Underbody Based on Deep Learning. IEEE Access 2019, 7, 183838–183846.